\documentclass[11pt]{article}

\usepackage[final]{acl}

\usepackage{times}
\usepackage{latexsym}

\usepackage[T1]{fontenc}
\usepackage[utf8]{inputenc}

\usepackage{microtype}

\usepackage{inconsolata}

\usepackage{graphicx}
\usepackage{booktabs}  % 为了 \toprule, \midrule, \bottomrule
\usepackage{multirow}  % 为了 \multirow
\usepackage{amssymb}
\usepackage{amsmath}   % 这是一个好习惯，通常都需要
\usepackage[many]{tcolorbox} % 导入 tcolorbox 包
\usepackage{xcolor} % 确保加载 xcolor
\usepackage{colortbl}

\title{Step-GRPO: Internalizing Dynamic Early Exit for Efficient Reasoning}

\author{
 \textbf{Benteng Chen}\textsuperscript{1,2,*} ,
 \textbf{Weida Wang}\textsuperscript{1,2,3,*,\S} ,
 \textbf{Shufei Zhang}\textsuperscript{2,$\dagger$} ,
 \textbf{Mingbao Lin}\textsuperscript{4} ,
 \textbf{Min Zhang}\textsuperscript{1,$\dagger$,$\ddagger$}
\\
 \textsuperscript{1}East China Normal University,
 \textsuperscript{2}Shanghai AI Laboratory, 
 \textsuperscript{3}Fudan University,
 \textsuperscript{4}Rakuten Singapore
\\
 \small \textsuperscript{*}Equal contribution \quad \textsuperscript{\S}Student project leader \quad \textsuperscript{$\ddagger$}Project leader 
\\
 \small \textsuperscript{$\dagger$}Corresponding authors: \href{mailto:mzhang@cs.ecnu.edu.cn}{mzhang@cs.ecnu.edu.cn}, \href{mailto:zhangshufei@pjlab.org.cn}{zhangshufei@pjlab.org.cn}
}

\tcbuselibrary{breakable}
\begin{document}
\maketitle
\begin{abstract}
%Large Reasoning Models (LRMs) leveraging Long Chain-of-Thought (Long-CoT) generation have demonstrated remarkable problem-solving capabilities but suffered from overthinking, generating redundant verification steps that inflate computational costs. Existing solutions to this inefficiency face a critical dilemma where training-time length penalties often induce syntactic blindness causing capability collapse, while inference-time early-exit mechanisms introduce substantial system overhead. 

Large reasoning models that use long chain-of-thought excel at problem-solving yet waste compute on redundant checks. Curbing this overthinking is hard: training-time length penalties can cripple ability, while inference-time early-exit adds system overhead.
To bridge this gap, we propose \emph{Step-GRPO}, a novel post-training framework that internalizes dynamic early-exit capabilities directly into the model. Step-GRPO shifts the optimization objective from raw tokens to semantic steps by utilizing linguistic markers to structure reasoning. We introduce a Dynamic Truncated Rollout mechanism that exposes the model to concise high-confidence trajectories during exploration, synergized with a Step-Aware Relative Reward that dynamically penalizes redundancy based on group-level baselines. Extensive experiments across three model sizes on diverse benchmarks demonstrate that Step-GRPO achieves a superior accuracy-efficiency trade-off. On Qwen3-8B, our method reduces token consumption by 32.0\% compared to the vanilla model while avoiding the accuracy degradation observed in traditional length-penalty methods. %This approach effectively yields a model with intrinsic efficiency and zero additional inference overhead.

\end{abstract}
\section{Introduction}
Large reasoning models (LRMs), like DeepSeek-R1~\citep{guo2025deepseek} and Qwen3~\citep{yang2025qwen3}, have well solved complex problems using long chain-of-thought (CoT). However, this reasoning ability comes at a high computational cost. A common problem is ``overthinking''~\citep{yang2025dynamic,dai2025stable}, models often generate unnecessary verification steps or circular explanations even after finding the correct solution.  This happens because standard reinforcement learning methods, like GRPO~\cite{shao2024deepseekmath}, primarily reward correct outcomes. Since longer reasoning chains statistically increase the chance of getting the right answer, the model naturally learns to be verbose, leading to wasted computation and higher latency.
\begin{figure}
\includegraphics[width=0.5\textwidth,height=\textheight,keepaspectratio]{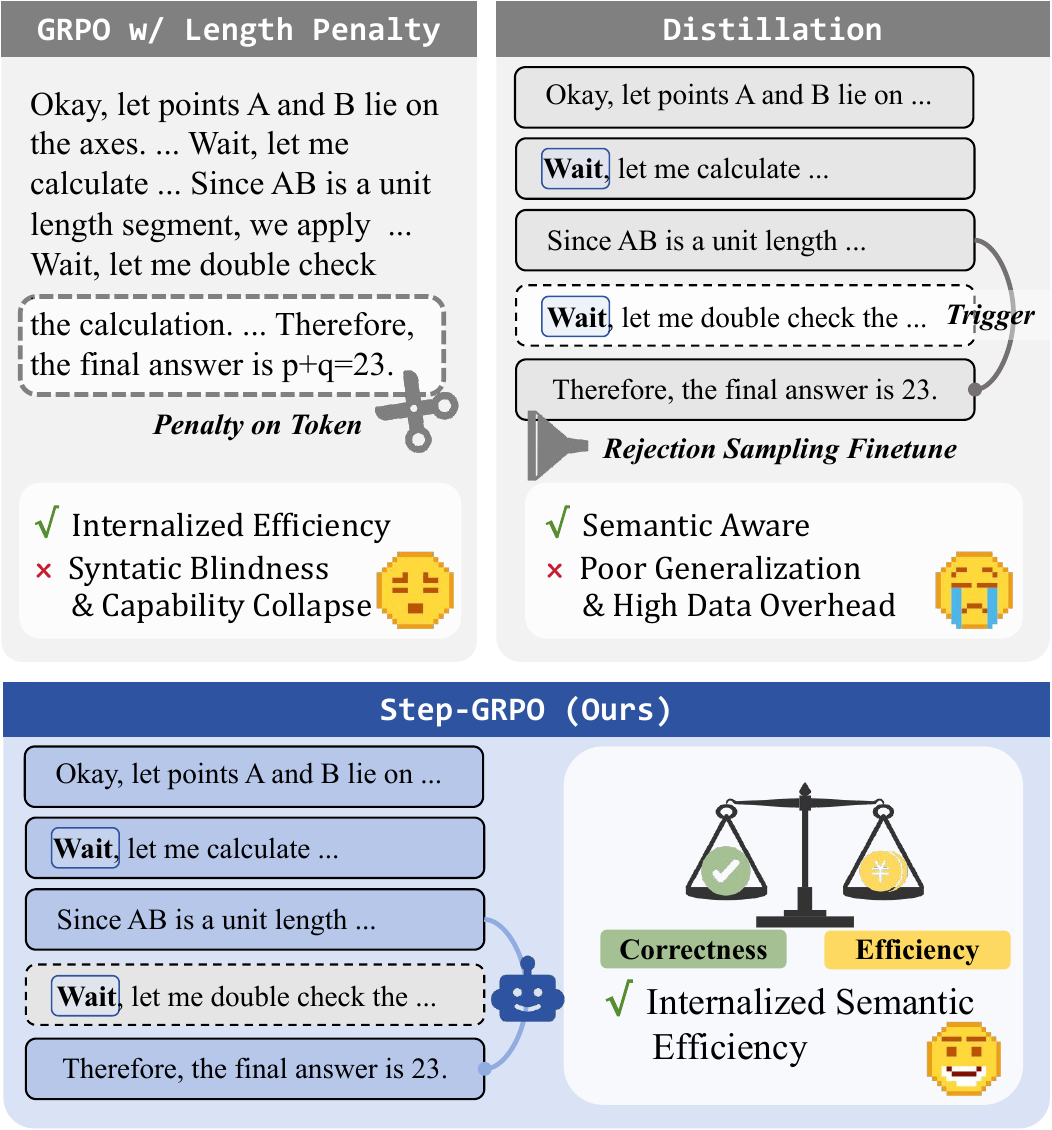} 
\caption{Advantages of our Step-GRPO.}
\vskip -0.25in 

\label{fig:intro}

\end{figure}

To address this inefficiency, recent research has focused on constraining the generation length during post-training~\cite{yu2025dapo,dai2025stable,team2025kimi15}.
% typically following two main directions. The first approach involves strict format constraints. Methods like standard GRPO often use format-based rewards or hard truncation~\cite{team2025kimi2} to force the model to adhere to a concise output structure. The second approach employs length-based reward shaping~\cite{yu2025dapo,dai2025stable,team2025kimi15}. For instance, DAPO~\cite{yu2025dapo} introduces an ``overlong reward shaping'' mechanism that uses soft length intervals to penalize excessive tokens. Similarly, recent works like GRPO with length penalty and GRPO-$\lambda$~\cite{dai2025stable} explicitly add a penalty term to the reward function to discourage long responses. 
Methods like GRPO with length penalty~\cite{team2025kimi15,yu2025dapo} or dynamic adjustments like GRPO-$\lambda$~\cite{dai2025stable} explicitly add penalty terms to the reward function.
% However, both directions share a major limitation: \emph{they rely on token counts, which creates a problem of ``syntactic blindness''.} The model cannot distinguish between necessary reasoning details and useless words. As a result, penalizing tokens often forces the model to cut important logic, leading to training instability or accuracy drops.

However, as illustrated in Figure~\ref{fig:intro} (Top Left), these approaches suffer from \emph{``syntactic blindness''}. Since they rely on indiscriminate token counting, models cannot distinguish between redundancy and necessary reasoning. Consequently, penalizing tokens often forces the model to cut essential verification steps, leading to \emph{capability collapse} where brevity is achieved at the cost of accuracy~\cite{gao2023scaling,su2025thinking}.

% Parallel research explores inference-time strategies using lexical cues~\citep{vanhoyweghen2025lexical,wang2025beyond,yang2025dynamic} or entropy~\citep{agarwal2025unreasonable} to optimize generation. While effective, they impose significant runtime overhead. To internalize this capability, recent works shift to supervised fine-tuning (SFT) on compressed trajectories~\citep{li2025compressing,ma2025cot,cui2025stepwise,qiao2025concise,zhang2025control}. However, this approach faces a data bottleneck, as constructing valid concise samples requires expensive rejection sampling. Critically, SFT yields \emph{superficial imitation}: the model mimics brevity but fails to generalize the underlying decision-making policy. This highlights the need for a framework that learns \emph{semantic efficiency} directly within the training loop, avoiding both ``syntactic blindness'' and costly data curation.
Another direction attempts to internalize efficient inference strategies~\citep{wang2025beyond,yang2025dynamic}, which rely on lexical cues~\citep{vanhoyweghen2025lexical} or entropy~\citep{agarwal2025unreasonable} to optimize generation, via distillation. 
Recent works apply supervised fine-tuning (SFT) on compressed trajectories~\citep{qiao2025concise,zhang2025control} derived from rejection sampling. 
While \emph{semantic-aware}, this approach faces severe limitations shown in Figure~\ref{fig:intro} (Top Right): constructing valid concise samples incurs \emph{high data overhead} due to expensive rejection sampling. Critically, SFT yields \emph{poor generalization}, the model superficially mimics the concise style of the training data without learning the underlying decision-making policy, often failing on complex unseen tasks.

To bridge this gap, we propose Step-GRPO, a novel post-training framework that \emph{internalizes the efficient reasoning capability} directly into the model itself (Figure~\ref{fig:intro}, Bottom). Unlike previous methods that focus on tokens, Step-GRPO addresses the problem through \emph{semantic steps}. 
First, we solve the syntactic blindness of previous training methods by moving from tokens to \emph{semantic steps}. We use linguistic markers to segment reasoning and apply a \emph{step-aware reward}. 
Second, we simulate the decision-making process of inference interventions during training rollouts. By mixing natural trajectories with teacher-guided truncated paths, we force the model to learn that fewer reasoning steps yield higher rewards. This effectively transfers the external stopping capability into the model itself, achieving semantically efficient reasoning with zero inference overhead.
Experimental results show that Step-GRPO achieves a superior trade-off between accuracy and efficiency, significantly reducing tokens while avoiding the training collapse in length-penalty methods.

\section{Related Work}

\subsection{Inference-Time Efficiency Interventions}
With the utilization of test-time scaling~\citep{snell2024scaling} to enhance complex tasks, LRMs~\citep{kojima2022large,jaech2024openai,guo2025deepseek} generate long CoT~\citep{lightman2023let} sequences, leading to increased computational load and inference latency. Research indicates an intrinsic ``overthinking'' phenomenon in LRMs, where models persistently generate verbose reasoning sequences with redundant steps, wasting resources and potentially degrading accuracy~\citep{chennot,saito2023verbosity}. To address this, existing work is two-fold: \emph{prompt-based} and \emph{output-based}. Prompt-based methods apply external constraints, such as NoThinking strategy~\citep{ma2025reasoning} that skips reasoning, Chain-of-Draft (CoD)~\citep{xu2025chain} which limits word counts, and Token-Conditional Control (TCC)~\citep{muennighoff2025s1} that sets token budgets. However, these static constraints often compromise reasoning capabilities or are ignored in complex tasks. Conversely, output-based methods aim for dynamic early exit. For example, Dynasor-CoT~\citep{fu2025reasoning} periodically checks answer consistency but often suffers from late termination. While methods like DEER~\citep{yang2025dynamic} achieve precise truncation, strategies requiring parallel decoding or branch evaluation often increase overhead, potentially negating efficiency gains.

\subsection{Training-Time Alignment for Conciseness}
To mitigate deployment latency during inference-time, research has internalized concise reasoning capabilities directly into model weights. SFT on compressed CoT data is a primary strategy~\citep{li2025compressing,ma2025cot,cui2025stepwise,xia2025tokenskip}. While approaches like ConCISE~\citep{ma2025cot,qiao2025concise,xia2025tokenskip,cui2025stepwise} improve efficiency via mixed training or confidence-guided data filtration, they face significant bottlenecks in constructing high-quality, logically complete yet concise samples. Alternatively, Reinforcement Learning (RL) targets efficiency via length penalties, yet often risks ``corner cutting'' and capability collapse~\citep{dai2025stable}. While recent innovations like GRPO-$\lambda$~\citep{dai2025stable}, entropy-guided compression~\citep{zhu2025entropy}, and S-GRPO~\citep{dai2025s} improve stability through dynamic adjustments or decaying rewards, they remain reliant on statistical heuristics (\emph{e.g.}, entropy, sequence length) that ignore reasoning semantics. Distinct from these scalar-based approaches, we integrate semantic stopping directly into the GRPO loop. By treating linguistic markers (\emph{e.g.}, ``Wait'') as decision anchors, we reward the model for logical closure at valid transition points, ensuring compression aligns with the natural structure of reasoning rather than blind truncation.
%integrity.

\begin{figure*}[!t]
    \centering
\includegraphics[width=\textwidth,height=\textheight,keepaspectratio]{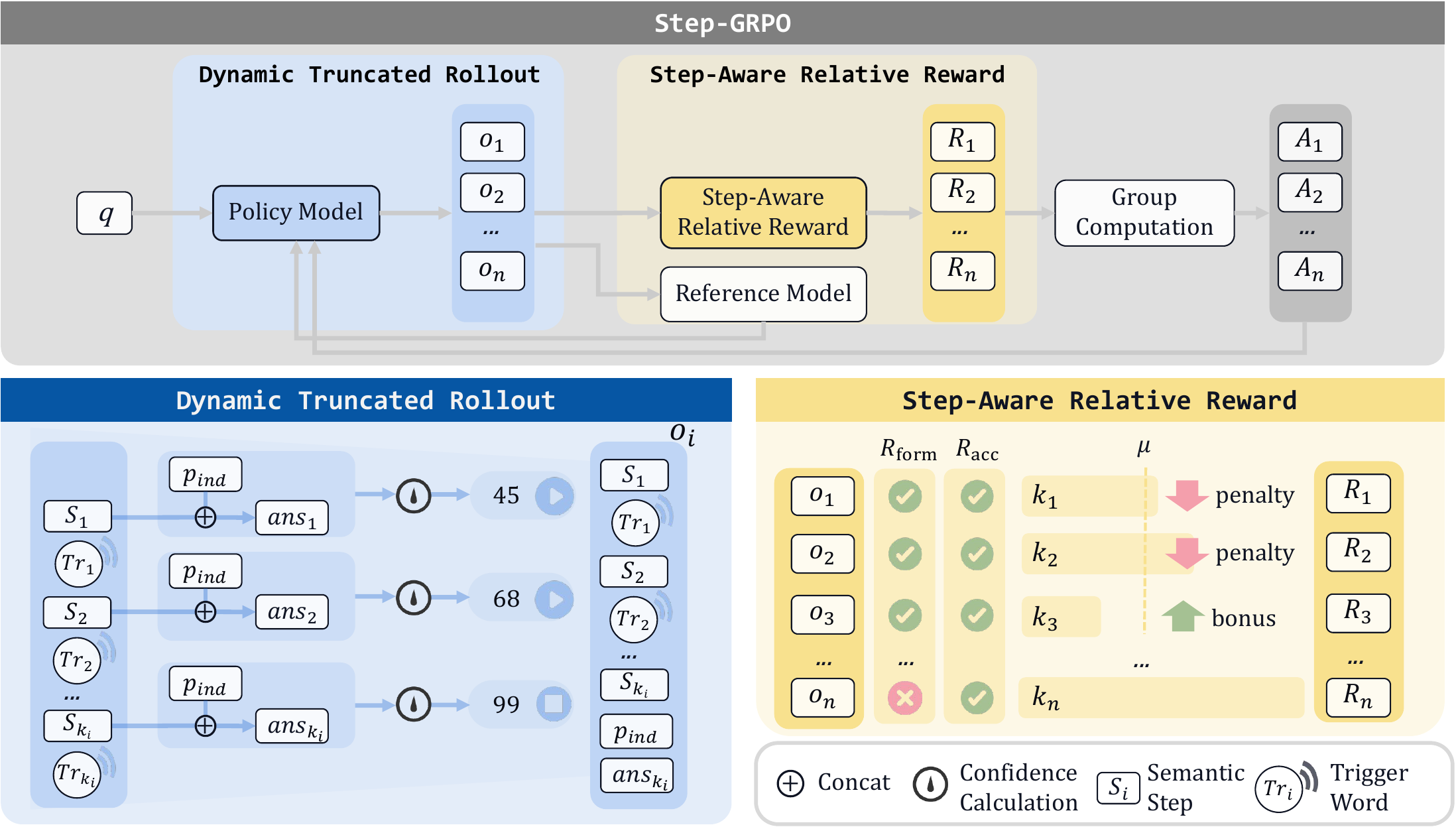} 
    \caption{The overall pipeline of Step-GRPO.}
    \label{fig:pipeline}
    % \vskip -0.2in
\end{figure*}

\section{Method}
In this section, we propose Step-GRPO, a post-training framework designed to internalize the dynamic early-exit capability into the reasoning model itself. As illustrated in Figure\,\ref{fig:pipeline}, Step-GRPO consists of three integral components: \emph{Dynamic Truncated Rollout} during exploration (Section \ref{subsec:rollout}), \emph{Semantic Step Quantification} (Section \ref{subsec:step}), and \emph{Step-Aware Relative Reward} (Section \ref{subsec:reward}) for policy optimization.

\subsection{Preliminary}
We consider a reasoning task where the input is a question $q$ and the ground truth is $y^*$. The policy model $\pi_\theta$, parameterized by $\theta$, generates a completion $o_i$, which consists of both the CoT reasoning path and the final answer. 
Following the GRPO framework~\cite{shao2024deepseekmath}, for each question $q$, we sample a group of $G$ outputs $\mathcal{G} = \{o_1, o_2, \dots, o_G\}$ from the old policy $\pi_{\theta_{\text{old}}}$. The objective is to maximize a reward  $J(\theta)$:
\begin{equation}
    J(\theta) = \mathbb{E}_{q \sim \mathcal{D}, o \sim \pi_\theta(\cdot|q)} [R(o, q)],
\end{equation}
where $R(o, q)$ is a reward function balancing correctness and reasoning efficiency.

\subsection{Dynamic Truncated Rollout}
\label{subsec:rollout}
To enable the model to learn efficient reasoning paths, we must expose it to \emph{short yet correct} trajectories during training. Instead of standard autoregressive generation, we enforce a \emph{Dynamic Truncated Rollout} mechanism for all samples in $\mathcal{G}$, inspired by inference-time early-exit strategies~\citep{yang2025dynamic,wang2025beyond}.
For each completion $o_i$, we utilize a set of transition \emph{trigger words} $\mathcal{W}_{trig}$ (\emph{e.g.}, ``Wait'', ``Alternatively''). Following \citet{qiao2025concise}, we define these triggers as signals for boundaries between \emph{semantic reasoning steps}. The generation proceeds iteratively as:

\paragraph{1) Semantic Step Detection}
We monitor the generation process continuously. Once a trigger word $Tr \in \mathcal{W}_{trig}$ is detected at the end of the current sequence, we define the content generated since the last trigger as a semantic step $S$. We then pause the standard generation to evaluate the necessity of further reasoning.

\paragraph{2) Answer Induction} 
We construct a temporary input context by appending an answer-inducing prompt $p_{ind}$ (\emph{e.g}., ``\textit{</think> The final answer is}'') to the current history. The model then generates a tentative answer $ans$.
    
\paragraph{3) Confidence Calculation} 
We evaluate the confidence $conf(ans)$ (we use $c(a)$ for short) of the tentative answer, defined as the average log-probability of its tokens:
    \begin{equation}
        c(a) = \frac{1}{|a|} \sum_{j=1}^{|a|} \log \pi_{\theta_{\text{old}}}(a_j \mid q, o_{i,<t}, p_{ind}, a_{<j}).
    \end{equation}
    
\paragraph{4) Truncation Decision} 
We compare the confidence against a threshold $\delta$.
    \begin{itemize}
        \item If $c(ans) > \delta$, the model is deemed confident. We terminate the reasoning process and define the final completion $o_i$ as the concatenation of the current path (including the current trigger $Tr$) and the induced answer $ans$.
        \item If $c(ans) \le \delta$, the tentative answer $ans$ is discarded. The model resumes generating the next semantic step until the next trigger word or the end of generation.
    \end{itemize}

This process ensures that the sampled group $\mathcal{G}$ contains diverse reasoning paths that are potentially truncated at the moment of sufficient confidence.

\subsection{Semantic Step Quantification}
\label{subsec:step}
Traditional efficiency metrics relying on raw token counts are often sensitive to phrasing verbosity. We instead evaluate computational cost through \emph{Semantic Steps}. 

We quantify the reasoning complexity $k_i$ for the $i$-th completion $o_i$ by tallying the occurrences of trigger words. Formally, let $N_{\text{trig}}(o_i)$ denote the total count of any trigger word $Tr \in \mathcal{W}_{trig}$ detected within the completion $o_i$. The semantic step count is defined as:
\begin{equation}
    k_i = 1 + N_{\text{trig}}(o_i),
\end{equation}
where the initial term $1$ accounts for the final reasoning segment (containing the answer) that typically follows the last trigger. This quantification aligns with the discrete decision points in the dynamic rollout, providing a robust, structure-aware metric for reasoning complexity.

\subsection{Step-Aware Relative Reward}
\label{subsec:reward}
We propose a \emph{Step-Aware Relative Reward} to guide the policy optimization. Unlike static length penalties, our reward mechanism employs a dynamic reference derived from the group's performance.

For each completion $o_i \in \mathcal{G}$, the total reward $R_i$ is computed based on its correctness and its relative step efficiency.

\paragraph{Dynamic Step Baseline}
We first calculate the dynamic average $\mu$, defined as the mean step count of all \emph{correct} completions in the current group $\mathcal{G}_{correct}$:
\begin{equation}
    \mu = \frac{1}{|\mathcal{G}_{correct}|} \sum_{j:o_j \in \mathcal{G}_{correct}} k_j.
\end{equation}

We exclude incorrect samples to prevent baseline skewing, as they often exhibit extreme step counts (\emph{e.g.}, premature guessing or circular hallucinations). Including such outliers would distort $\mu$, leading to misaligned efficiency incentives. If the group contains no correct answers, we omit efficiency term.

\paragraph{Final Reward Function}
The total reward $R_i$ serves as a composite objective, balancing solution correctness, reasoning efficiency, and structural compliance. It combines an accuracy indicator $R_{\text{acc}}^{(i)}$, a step-aware efficiency term, and a format consistency reward $R_{\text{form}}^{(i)}$:
\begin{equation}
\begin{aligned}
    R_i = &\alpha \cdot R_{\text{acc}}^{(i)} \cdot \left[ 1 - \beta \cdot \tanh \left( \frac{k_i-\mu}{\mu} \right) \right] \\
    &+(1- \alpha) \cdot R_{\text{form}}^{(i)},
\end{aligned}
\end{equation}
where $\alpha \in [0,1]$ balances accuracy against formatting constraints, and $\beta > 0$ controls the penalty strength. 
The term $(\frac{k_i - \mu}{\mu})$ represents the relative deviation from the group mean. By applying the hyperbolic tangent function ($\tanh$), we bound the \emph{step efficiency incentive} to the range $(-\beta, \beta)$. This formulation prevents extreme variance in rewards while encouraging the model to seek the \emph{minimal sufficient reasoning path}. Specifically, if a correct response uses fewer steps than average ($k_i < \mu$), the penalty term becomes a bonus (positive), increasing $R_i$ above $\alpha$; otherwise, it acts as a penalty.

\subsection{Policy Optimization}
Following the GRPO framework~\citep{shao2024deepseekmath}, we optimize the policy $\theta$ by maximizing the expected advantage over the generated tokens. First, we compute the advantage $A_i$ for each completion $o_i$ by standardizing the rewards within the group:
\begin{equation}
    A_i = \frac{R_i - \text{mean}(\{R_1, \dots, R_G\})}{\text{std}(\{R_1, \dots, R_G\})}.
\end{equation}

The final objective function is defined as the average per-token importance-weighted advantage, constrained by a KL divergence term. To maintain readability, we use $\pi(\cdot)$ to denote the conditional dependence $\pi(o_{i,t}|q, o_{i,<t})$:
\begin{equation}
\begin{aligned}
J_{\text{GRPO}}(\theta) = \mathbb{E} & \left[ \frac{1}{G} \sum_{i=1}^G \sum_{t=1}^{|{o}_i|} 
\min \left( \frac{\pi_{\theta}(\cdot)}{\pi_{\theta_{\text{old}}}(\cdot)} A_i, \right. \right. \\
& \quad \left. \left. \text{clip} \left( \frac{\pi_{\theta}(\cdot)}{\pi_{\theta_{\text{old}}}(\cdot)}, 1-\epsilon, 1+\epsilon \right) A_i \right) \right. \\
& \quad \left. - \beta_{\text{KL}} D_{\text{KL}}(\pi_\theta \| \pi_{\text{ref}}) \vphantom{\sum_{i=1}^G} \right].
\end{aligned}
\end{equation}
where $\epsilon$ is the clipping parameter used in Proximal Policy Optimization (PPO) to limit policy updates~\citep{schulman2017proximal}.

\section{Experiment}
\subsection{Experimental Setup}

\paragraph{Benchmarks}
% GSM8K, GPQA, MATH-500, AMC 2023, AIME 2024
To comprehensively evaluate the accuracy-efficiency trade-off, we conduct experiments across diverse benchmarks spanning varying difficulty levels. We utilize GSM8K~\citep{cobbe2021training} and MATH-500~\citep{hendrycks2021measuring} for standard multi-step reasoning tasks. To assess performance on complex problems requiring extensive reasoning chains, we include AMC 2023~\cite{aimo2024amc}, AIME 2024~\cite{aime24}, and AIME 2025~\citep{aime25}. Additionally, we test domain-specific expert reasoning using GPQA~\citep{rein2024gpqa} (Diamond subset).

\paragraph{Baselines}
% vanilla GRPO, GRPO+ length clip, GRPO + length penalty, GRPO-$\lambda$, S-GRPO, DEER
We compare our method against a comprehensive set of baselines to evaluate both reasoning accuracy and generation efficiency. \emph{Vanilla} denotes the original base model without additional reinforcement learning. For standard RL comparisons, \emph{GRPO} represents the original Group Relative Policy Optimization algorithm trained purely on correctness rewards. To benchmark efficiency-oriented strategies, we include \emph{GRPO-8k}, which imposes a hard truncation at 8,192 tokens during training to simulate resource-constrained generation. We further compare against three state-of-the-art penalty-based methods: \emph{GRPO+LP}, which adopts the length penalty mechanism implemented in Kimi 1.5~\citep{team2025kimi15}; \emph{GRPO+SOP}, which utilizes the Soft Overlong Punishment introduced in DAPO~\citep{yu2025dapo}; and \emph{GRPO-$\lambda$}~\citep{dai2025stable}, which dynamically applies length penalties based on the correctness ratio of the rollout group. Finally, to assess the necessity of reinforcement learning over supervised distillation, we include \emph{DEER+SFT}, where the model is finetuned on concise and correct reasoning chains collected via DEER~\citep{yang2025dynamic}-guided rejection sampling.

\paragraph{Implementation Details}
Our implementation is built upon the EasyR1 training framework~\citep{zheng2025easyr1}. For training, we utilize data selected from the DAPO-Math-17k dataset~\citep{yu2025dapo}. We define the transition trigger tokens $\mathcal{W}_{trig}$ following \citet{qiao2025concise}, including terms such as ``Wait'' and ``Alternatively''. Regarding hyperparameters, we set the reward balancing coefficient $\alpha=0.1$ step penalty strength $\beta = 0.5$ and the rollout group size $G=5$. During the dynamic truncated rollout, the confidence threshold $\delta$ is set to 0.95. For all RL experiments, we set the global batch size to 512, use a constant learning rate of $1\text{e}-6$ and a KL penalty coefficient $\beta_{KL} = 0.01$. The maximum generation length is set to 16,384 tokens for all models (except \emph{GRPO-8k}) to allow sufficient reasoning exploration. All experiments are conducted on $8 \times \text{H100}$ GPUs.

\paragraph{Evaluation Metrics}
In addition to accuracy, we evaluate reasoning efficiency using the Compression Rate (CR). To faithfully reflect the model's balanced performance across tasks of different difficulty, we compute CR per task first, then take the arithmetic mean:
\begin{equation}
    CR_{\text{overall}} = \frac{1}{|\mathcal{D}|} \sum_{i \in \mathcal{D}} \frac{\text{Avg\_Tok}_{\text{Model}}^{(i)}}{\text{Avg\_Tok}_{\text{Vanilla}}^{(i)}}
\end{equation}
where $\mathcal{D}$ denotes the set of evaluated benchmark datasets. Lower values indicate better compression.

\subsection{Main Results}
\begin{table*}[t]
\centering
% 定义自定义颜色 (R, G, B) = (221, 233, 247)
\definecolor{oursblue}{RGB}{221,233,247}

\caption{Experimental results. The order is arranged as requested. ``Acc'' denotes accuracy, ``Tok'' denotes token count, and ``CR'' denotes compression rate (relative to Vanilla). Intermediate CR columns are omitted for brevity. Best results are in \textbf{bold}, and second-best results are \underline{underlined}.}
\resizebox{\textwidth}{!}{
\begin{tabular}{l|cc|cc|cc|cc|cc|cc|cc}
\toprule
\multirow{2}{*}{\textbf{Method}} & \multicolumn{2}{c}{\textbf{GSM8K}} & \multicolumn{2}{c}{\textbf{GPQA}} & \multicolumn{2}{c}{\textbf{MATH-500}} & \multicolumn{2}{c}{\textbf{AMC 2023}} & \multicolumn{2}{c}{\textbf{AIME 2024}} & \multicolumn{2}{c}{\textbf{AIME 2025}} & \multicolumn{2}{c}{\textbf{Overall}} \\
& Acc $\uparrow$ & Tok $\downarrow$ & Acc $\uparrow$ & Tok $\downarrow$ & Acc $\uparrow$ & Tok $\downarrow$ & Acc $\uparrow$ & Tok $\downarrow$ & Acc $\uparrow$ & Tok $\downarrow$ & Acc $\uparrow$ & Tok $\downarrow$ & Acc $\uparrow$ & CR $\downarrow$ \\
\midrule
\rowcolor{gray!20} \multicolumn{15}{c}{\textit{Qwen3-8B}} \\
Vanilla & 93.2 & 2,362 & \underline{62.1} & 7,249 & \textbf{96.8} & 5,299 & \textbf{97.5} & 8,648 & 70.0 & 15,880 & 60.0 & 17,492 & 79.9 & 100\% \\
DEER+SFT & 90.5 & 1,733 & 47.0 & 8,860 & 95.0 & 3,361 & 90.0 & 6,149 & 63.3 & 10,623 & 50.0 & 13,330 & 72.6 & 78.9\% \\
GRPO & 93.3 & 2,262 & 54.6 & 6,087 & \textbf{96.8} & 5,092 & \textbf{97.5} & 8,204 & \textbf{76.7} & 12,545 & \underline{66.7} & 15,407 & \underline{80.9} & 89.7\% \\
GRPO-8k & \underline{94.2} & 1,754 & 55.1 & 5,085 & 96.2 & 3,796 & 87.5 & 5,933 & 66.7 & 10,422 & 53.3 & 12,638 & 75.5 & 70.4\% \\
GRPO+LP & \underline{94.2} & \textbf{813} & 55.6 & \textbf{4,820} & 95.0 & \textbf{2,336} & 92.5 & \textbf{3,786} & \underline{73.3} & \textbf{9,899} & 60.0 & \underline{11,932} & 78.4 & \textbf{53.2\%} \\
GRPO+SOP & 93.0 & 1,582 & \textbf{62.6} & 5,330 & 95.8 & 3,822 & \textbf{97.5} & 5,431 & 66.7 & \underline{10,173} & \underline{66.7} & \textbf{11,844} & 80.4 & 67.9\% \\
GRPO-$\lambda$ & 92.9 & \underline{1,193} & 60.1 & \underline{5,003} & \underline{96.4} & \underline{3,095} & 90.0 & 5,432 & \textbf{76.7} & 10,554 & 63.3 & 12,321 & 79.9 & \underline{62.9\%} \\
\rowcolor{oursblue} \textbf{Step-GRPO (Ours)} & \textbf{94.9} & 1,354 & 56.1 & 5,560 & \textbf{96.8} & 3,485 & \underline{95.0} & \underline{5,295} & \textbf{76.7} & 11,657 & \textbf{73.3} & 12,859 & \textbf{82.1} & 68.0\% \\
\midrule
\rowcolor{gray!20} \multicolumn{15}{c}{\textit{Qwen3-4B}} \\
Vanilla & \underline{93.8} & 2,242 & \textbf{55.1} & 6,644 & 96.2 & 5,346 & \textbf{97.5} & 8,801 & 70.0 & 15,453 & \textbf{63.3} & 18,049 & \textbf{79.3} & 100\% \\
DEER+SFT & 90.1 & 1,869 & 37.4 & 8,404 & 93.0 & 3,610 & 85.0 & 5,002 & 53.3 & 12,105 & 46.7 & 12,503 & 67.6 & 80.3\% \\
GRPO & 93.0 & 2,291 & 51.0 & 5,380 & \textbf{97.6} & 4,974 & \textbf{97.5} & 7,712 & \underline{73.3} & 12,126 & \textbf{63.3} & 14,249 & \textbf{79.3} & 86.9\% \\
GRPO-8k & \underline{93.8} & 1,753 & 50.5 & 4,571 & 96.2 & 3,691 & \underline{95.0} & 5,506 & 70.0 & 10,515 & \underline{60.0} & \underline{11,172} & 77.6 & 68.1\% \\
GRPO+LP & 93.0 & \textbf{680} & 50.0 & \textbf{4,039} & 94.2 & \textbf{2,329} & \underline{95.0} & \underline{4,712} & 70.0 & 10,631 & \textbf{63.3} & \textbf{10,783} & 77.6 & \textbf{52.8\%} \\
GRPO+SOP & \textbf{93.9} & 1,656 & 50.0 & \underline{4,452} & 95.6 & 3,808 & \underline{95.0} & 5,793 & 70.0 & \underline{9,363} & \underline{60.0} & 13,838 & 77.4 & 69.2\% \\
GRPO-$\lambda$ & 93.1 & \underline{1,261} & 48.0 & 4,606 & \underline{96.6} & \underline{3,232} & \textbf{97.5} & \textbf{4,671} & 70.0 & \textbf{9,094} & \underline{60.0} & 12,814 & 77.5 & \underline{61.5\%} \\
\rowcolor{oursblue} \textbf{Step-GRPO (Ours)} & 93.3 & 1,298 & \underline{54.0} & 5,182 & 95.6 & 3,528 & \underline{95.0} & 5,456 & \textbf{76.7} & 11,435 & \underline{60.0} & 13,356 & \underline{79.1} & 68.7\% \\
\midrule
\rowcolor{gray!20} \multicolumn{15}{c}{\textit{Qwen3-1.7B}} \\
Vanilla & \underline{89.2} & 2,326 & 37.4 & 6,781 & \underline{90.8} & 5,777 & 80.0 & 10,713 & 43.3 & 18,552 & 26.7 & 17,766 & 61.2 & 100\% \\
DEER+SFT & 84.2 & 2,811 & 22.2 & 14,165 & 83.0 & 4,615 & 67.5 & \textbf{5,712} & 30.0 & \underline{12,758} & 20.0 & \underline{12,367} & 51.2 & 100.2\% \\
GRPO & \textbf{90.1} & 2,100 & \textbf{39.4} & 6,037 & \textbf{92.4} & 5,140 & \textbf{85.0} & 9,425 & \underline{50.0} & 15,039 & \underline{40.0} & 16,427 & \textbf{66.1} & 88.3\% \\
GRPO-8k & 88.9 & 1,688 & \underline{38.9} & 5,123 & 90.6 & 3,931 & 80.0 & 7,021 & 43.3 & 13,600 & 33.3 & 12,362 & 62.5 & 70.8\% \\
GRPO+LP & \textbf{90.1} & \textbf{951} & \underline{38.9} & \textbf{4,691} & 89.4 & \textbf{2,718} & 72.5 & \underline{6,605} & \textbf{53.3} & \textbf{11,508} & 20.0 & \textbf{11,412} & 60.7 & \textbf{57.5\%} \\
GRPO+SOP & \underline{89.2} & 1,778 & 36.9 & 5,188 & 90.2 & 4,419 & \underline{82.5} & 7,704 & 40.0 & 13,934 & 30.0 & 14,216 & 61.5 & 76.1\% \\
GRPO-$\lambda$ & 88.7 & 1,431 & 34.3 & \underline{4,899} & 90.2 & \underline{3,708} & 75.0 & 7,050 & 40.0 & 13,572 & \textbf{43.3} & 12,994 & 61.9 & \underline{68.3\%} \\
\rowcolor{oursblue} \textbf{Step-GRPO (Ours)} & 88.5 & \underline{1,418} & 34.9 & 5,356 & 90.2 & 3,812 & \underline{82.5} & 6,856 & 43.3 & 13,728 & \underline{40.0} & 13,146 & \underline{63.2} & 69.7\% \\
\bottomrule
\end{tabular}
}
\label{tab:main_results}
\end{table*}  

Table~\ref{tab:main_results} presents the performance of Step-GRPO compared to baselines across three model scales. Our empirical results suggest that Step-GRPO significantly improves the trade-off boundary between accuracy and reasoning efficiency, outperforming baselines in this region.

\paragraph{Superior Accuracy-Efficiency Trade-off}
On Qwen3-8B, Step-GRPO achieves 82.1\% overall accuracy, surpassing the vanilla model while reducing token usage by 32.0\%. Although the aggressive GRPO+LP baseline yields higher compression, it severely harms reasoning performance. Notably, on the hard AIME 2025 benchmark, GRPO+LP accuracy drops to 60.0\%, showing that static token penalties force the model to cut essential reasoning steps. In contrast, Step-GRPO maintains 73.3\% accuracy on the same task. This proves that our step-aware reward correctly distinguishes between redundancy and necessary logical complexity, preserving the model's ability to reason deeply.

\paragraph{Necessity of Reinforcement Learning}
The results expose the limitations of supervised distillation. Creating the DEER+SFT dataset requires costly rejection sampling. Despite this effort, the resulting model generalizes poorly. On smaller models (4B and 1.7B), DEER+SFT suffers from severe instability on out-of-distribution (OOD) tasks. For instance, on the scientific reasoning benchmark GPQA, it exhibits negative compression (CR > 120-200\%), generating hallucinations significantly longer than Vanilla while suffering drastic accuracy drops This suggests SFT only mimics surface-level conciseness within the training distribution rather than internalizing the generalizable logic of when to stop. While other RL baselines like GRPO+SOP and GRPO-$\lambda$ improve over static penalties, they still lag behind Step-GRPO on complex tasks, confirming the advantage of our semantic step quantification.

\paragraph{Consistency Across Model Scales}
Experiments on Qwen3-4B and 1.7B confirm our method's robustness. Smaller models generally possess fragile reasoning capabilities. While GRPO+LP causes accuracy degradation (\emph{e.g.}, AIME 2025 drops to 20.0\% on 1.7B), Step-GRPO maintains a competitive accuracy profile (\emph{e.g.}, 40.0\% on AIME 2025) comparable to or better than standard baselines, while achieving over 30\% compression. This robustness stems from our dynamic group-relative baseline, which adapts the difficulty to the model's current capability. Furthermore, Step-GRPO effectively targets the repetitive loops common in smaller models, achieving high compression on simple tasks without the catastrophic collapse seen in static penalty methods.

\section{Discussion}
\subsection{Ablation Study} 

To validate the contribution of each component in Step-GRPO, we conduct ablation studies on Qwen3-8B using three representative benchmarks. The results are summarized in Table~\ref{tab:ablation}.

\begin{table}[h]
\centering
\caption{Ablation studies on Qwen3-8B. We report Accuracy (\%) and Average Token Count. The values demonstrate that removing the step reward leads to verbosity, while using a noisy all-sample baseline causes the worst performance degradation.}
\label{tab:ablation}
\resizebox{\linewidth}{!}{
\begin{tabular}{l|cc|cc|cc|cc}
\toprule
\multirow{2}{*}{\textbf{Method}} & \multicolumn{2}{c|}{\textbf{GPQA}} & \multicolumn{2}{c|}{\textbf{MATH-500}} & \multicolumn{2}{c|}{\textbf{AIME 2024}} & \multicolumn{2}{c}{\textbf{Average}} \\
 & Acc & Tok & Acc & Tok & Acc & Tok & Acc & Tok \\
\midrule
\textbf{Step-GRPO (Ours)} & 56.1 & 5560 & \textbf{96.8} & \textbf{3485} & \textbf{76.7} & 11657 & \textbf{76.5} & 6901 \\
\midrule
w/o Step Reward & 56.1 & 6300 & 96.4 & 4737 & 73.3 & 12787 & 75.3 & 7941 \\
w/o Dynamic Rollout & \textbf{56.6} & \textbf{5319} & 95.8 & 3822 & 70.0 & \textbf{10173} & 74.1 & \textbf{6438} \\
w/ All-Sample Mean & 54.6 & 5620 & 96.6 & 3994 & 66.7 & 12250 & 72.6 & 7288 \\
\bottomrule
\end{tabular}
}
\end{table}

\paragraph{Impact of Step-Aware Relative Reward}
To empirically verify whether Step-GRPO overcomes the \emph{syntactic blindness} of traditional length penalties, we designed a fine-grained structural analysis using GPT-4o to dissect the reasoning traces of 100 random samples. We explicitly categorized steps into \emph{Forward Reasoning}, \emph{Verification}, and \emph{Redundancy}. This allows us to distinguish between the \emph{essential verification} required for complex tasks and the \emph{harmful overthinking} that inflates computation. As shown in Table~\ref{tab:ablation}, removing this component (w/o Step Reward) leads to a significant rebound in token usage (from 6901 to 7941), confirming that without explicit semantic regularization, the model inevitably drifts back to the verbose generation patterns inherent in the original policy.

\begin{figure*}[t]
    \centering
\includegraphics[width=\textwidth,height=\textheight,keepaspectratio]{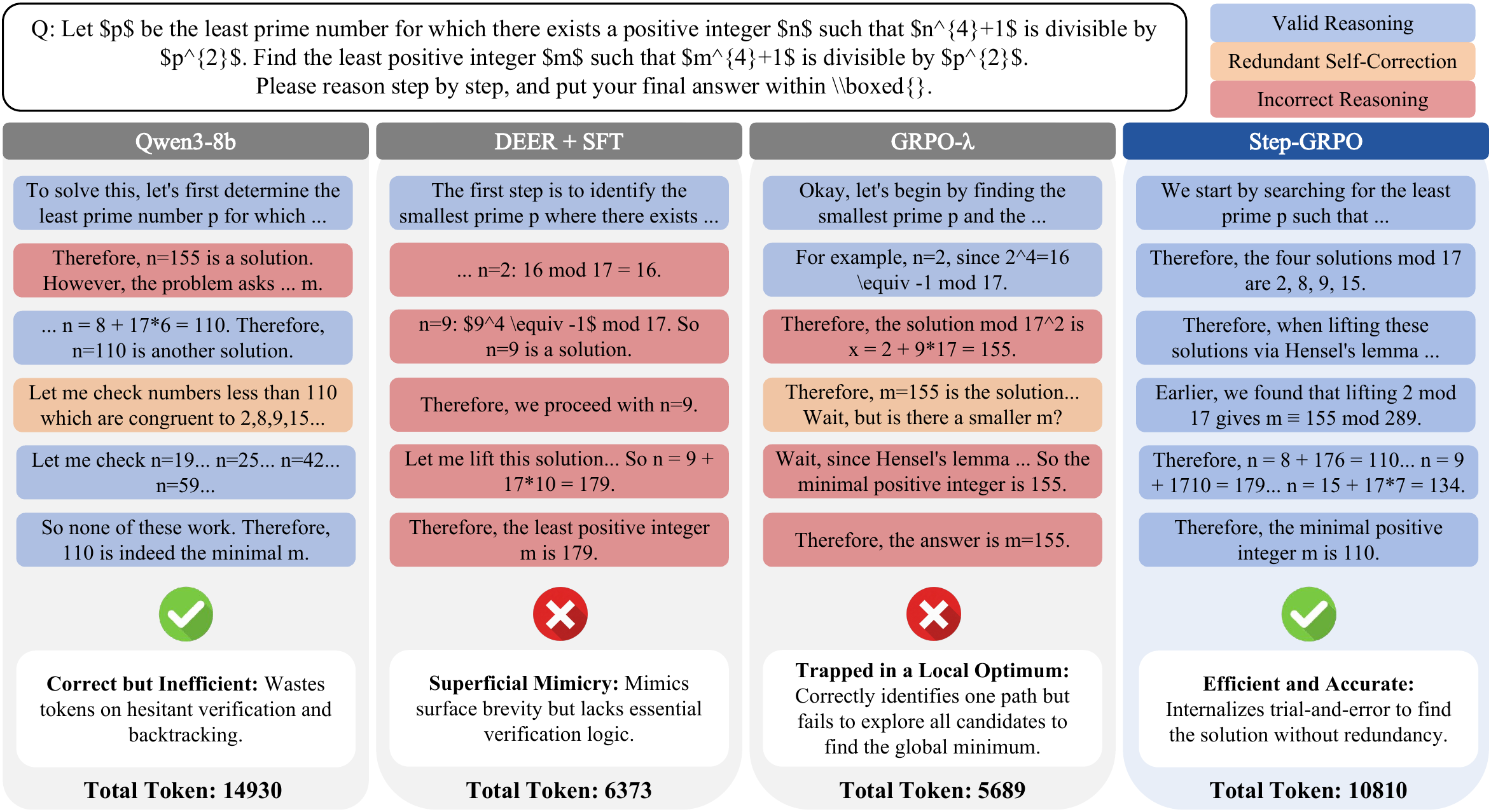} 
    \caption{Qualitative comparison of reasoning chains on a number theory problem from AIME 2024.}
    \label{fig:case_study}
    % \vskip -0.2in
\end{figure*}

\paragraph{Impact of Dynamic Truncated Rollout}
The variant without dynamic rollout (Row 3 in Table~\ref{tab:ablation}) shows a reduction in token usage due to the presence of the step penalty, but suffers a drop in accuracy (Avg 74.1\%). Without exposure to ``short yet correct'' trajectories during training, the model \emph{struggles to internalize the correct stopping logic}, forcing brevity at the cost of reasoning depth on complex tasks like AIME. This performance gap highlights a critical distribution mismatch: without the dynamic rollout acting as a rehearsal mechanism, the model treats early exit signals as out-of-distribution events, leading to hesitant and incomplete reasoning chains during inference.

\paragraph{Importance of Robust Baseline Calculation}
Calculating the baseline $\mu$ using all samples (Row 4 in Table~\ref{tab:ablation}) yields the lowest overall accuracy (72.6\%). Including incorrect responses, which are often extremely short (give-up) or long (circular), introduces severe noise into the reward signal. For instance, short but incorrect ``give-up'' responses artificially lower the group mean $\mu$, which inadvertently causes the reward function to penalize necessary, high-quality long reasoning steps as ``inefficient''. This \emph{confused reward baseline} prevents the model from effectively distinguishing between efficient reasoning and failure, leading to suboptimal convergence.

\subsection{Case Study}

\label{subsec:case_study}

To visualize the behavioral shift induced by our method, Figure~\ref{fig:case_study} presents a qualitative comparison on a complex number theory problem requiring strict constraint satisfaction. The Vanilla model, while correct, exhibits a typical trial-and-error pattern characterized by \emph{redundant hesitation} and backtracking, significantly inflating the computational cost. In contrast, optimization baselines like DEER+SFT and GRPO-$\lambda$ succumb to \emph{reasoning collapse}; they achieve brevity through a superficial mimicry of concise forms but fail to execute the critical checks required for this specific task, resulting in logical errors. Crucially, Step-GRPO demonstrates a successful \emph{internalization} of the decisive reasoning policy. It effectively prunes the self-doubting loops observed in the Vanilla model but, unlike the aggressive baselines, preserves the \emph{essential verification steps} necessary to identify the ``least positive'' integer, thereby achieving the correct solution with superior structural efficiency.

\begin{figure*}[t]
    \centering
    \includegraphics[width=\textwidth]{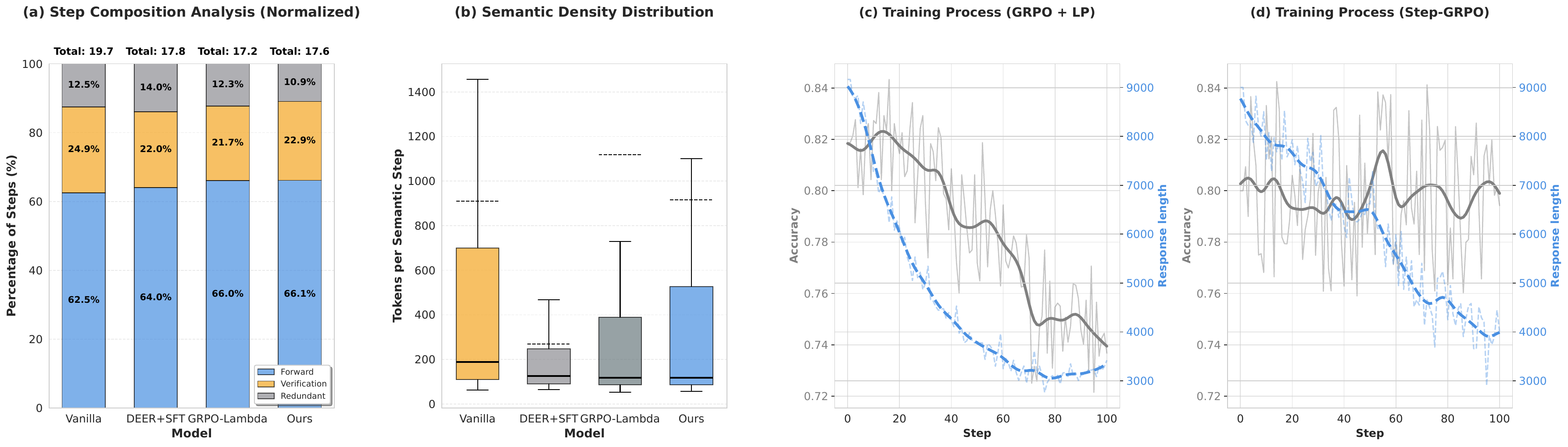}
\caption{\textbf{Structural and Training Dynamics Analysis.} 
\textbf{(a) Step Composition Analysis.} Proportions of step types with average step counts annotated on top.
\textbf{(b) Semantic Density Distribution.} Tokens per step (outliers excluded); dashed lines denote means.
\textbf{(c)(d) Training Dynamics.} Evolution of accuracy (Grey) and length (Blue) for GRPO+LP and Step-GRPO.}
    \label{fig:analysis}
    \vskip -0.1in
\end{figure*}

\subsection{Why Semantic Steps Matter?}

To empirically verify whether Step-GRPO overcomes the \emph{syntactic blindness} of traditional length penalties, we designed a fine-grained structural analysis using GPT-4o to dissect the reasoning traces of 200 random samples. We explicitly categorized steps into \emph{Forward Reasoning}, \emph{Verification}, and \emph{Redundancy}. This allows us to distinguish between the \emph{essential verification} required for complex tasks and the \emph{harmful overthinking} that inflates computation.

\paragraph{Selective Pruning of Redundancy.}
Our analysis reveals a critical divergence in how models achieve brevity. As shown in Figure~\ref{fig:analysis}(a), while token-based penalties (e.g., GRPO-$\lambda$) compress reasoning by indiscriminately suppressing all step types, Step-GRPO exhibits a selective pruning behavior. It reduces \emph{redundant} steps (Grey bars) to the lowest level among all models (10.9\%), yet retains a higher proportion of \emph{verification} steps (22.9\%, Yellow bars) compared to GRPO-$\lambda$ (21.7\%). This confirms that our method decouples reasoning efficiency from raw length, internalizing a policy that cuts \emph{syntactic fat} (verbosity) while preserving \emph{cognitive muscle} (verification).

\paragraph{Stabilizing Information Density.}
Beyond composition, we analyze the semantic density in Figure~\ref{fig:analysis}(b). This metric reflects the information payload per reasoning unit, quantified as:
\begin{equation}
    \text{Semantic Density}_i = \frac{\text{Semantic Step Count } k_i}{\text{Total Tokens in } o_i}
\end{equation}
The Vanilla model (Orange box) exhibits extreme variance with a ``tall'' distribution, indicating unpredictable ``overthinking'' loops where the model generates excessive tokens with low information gain. In contrast, Step-GRPO (Blue box) significantly compresses this variance. Unlike GRPO-$\lambda$ which often forces premature truncation (yielding a lower median density), our method maintains a compact and consistent density distribution. This suggests that Step-GRPO stabilizes the reasoning process, avoiding the fragmentation observed in static length-penalty methods by effectively filtering out ``low-density'' redundant steps.

\subsection{Training Stability}

% \begin{figure}
%     \framebox[0.48\textwidth]{\vbox to 5cm {}} % 创建一个指定宽高的空线框
% \caption{Training Stability.}
% \label{fig:stability}
% \end{figure}

To investigate the robustness of our method against the ``training collapse'' phenomenon, we visualize the evolution of response length and accuracy on the validate dataset in Figure~\ref{fig:analysis}(c)(d). 
The results reveal a critical failure mode in static token penalties (Left): \textit{GRPO+LP} shows a strong correlation between the rapid decrease in length (Blue dashed line) and a \emph{simultaneous, precipitous drop in accuracy} (Gray solid line, from $\sim$0.82 to $\sim$0.74). This indicates that the model is forced to indiscriminately abandon essential reasoning steps to satisfy the fixed penalty. 
Conversely, \emph{Step-GRPO} (Right) exhibits a \emph{decoupled trajectory}: while the reasoning length is significantly compressed (from $\sim$9k to $\sim$4k tokens), the accuracy remains robustly stable around 0.82. This demonstrates that our dynamic, group-relative baseline effectively acts as an \emph{adaptive curriculum}, penalizing only relative redundancy while \emph{protecting the necessary cognitive depth} required for complex problem-solving.

\section{Conclusion}
In this paper, we introduced \emph{Step-GRPO}, a novel post-training framework that enables Large Language Models to internalize efficient reasoning strategies. By synergizing dynamic truncated rollouts with a semantic step-aware relative reward, our method shifts the optimization objective from syntactic token minimization to semantic logic condensation. Extensive experiments across diverse benchmarks demonstrate that Step-GRPO achieves a superior accuracy-efficiency trade-off, significantly reducing computational costs while maintaining robustness on complex reasoning tasks. Crucially, we show that our dynamic, group-relative mechanism effectively resolves the ``training collapse'' issue plaguing traditional length-penalty methods. Ultimately, this work offers a scalable path toward efficient reasoning: transforming the explicit ``early-exit'' decision-making process into the model's intrinsic intuition, yielding decisive and precise reasoning chains without the need for external inference-time interventions.

\section*{Limitation}
\label{sec:limitations}

While Step-GRPO successfully internalizes efficient reasoning, the \emph{Dynamic Truncated Rollout} mechanism introduces a marginal increase in training latency during the generation phase, primarily due to the additional forward passes for confidence estimation. However, this overhead is partially mitigated by the accelerated parameter update phase, which benefits from significantly shorter sequence lengths. We consider this a justifiable ``training-time investment'' as it yields a zero-overhead model for deployment. Additionally, our current semantic step quantification relies on explicit linguistic markers, which may limit applicability in domains lacking such structures. Future work will focus on developing domain-agnostic step segmentation methods and exploring iterative self-training to reduce reliance on predefined triggers.

\section*{Ethics Statement}
This research complies with the ARR Ethics Policy. The datasets used in this study (e.g., GSM8K, MATH-500, AIME) are established, publicly available benchmarks for mathematical and logical reasoning, ensuring no violation of privacy or copyright. No human subjects or crowdworkers were employed in this research. 
We believe this work presents no significant risk of harm and offers a positive societal impact by democratizing efficient reasoning.

\section*{Acknowledgements}
This work was supported by the Key Laboratory of Cognitive Intelligence and Content Security, Ministry of Education (Grant No.10120251107, Harbin Institute of Technology). The National Natural Science Foundation of China (Grant No. 62477012), the AI for Science Program of the
Shanghai Municipal Commission of Economy and Informatization, China (Grant No. 2025-GZL-RGZN-BTBX-01014) and the robotic AI-Scientist platform of Chinese Academy of Sciences.

\bibliography{References}
\definecolor{example_back}{rgb}{1, 1, 1}
\definecolor{example_front}{rgb}{0.781, 0.874, 0.945}
\definecolor{prompt_front}{rgb}{0.859, 0.918, 0.922}
\definecolor{prompt_back}{rgb}{0.953, 0.971, 0.976}
\definecolor{human_front}{rgb}{0.784, 0.686, 0.604}
\definecolor{human_back}{rgb}{0.976, 0.898, 0.859}
\definecolor{example_front}{rgb}{0.8, 0.8, 0.8}
\definecolor{case_back}{rgb}{0.941, 0.941, 0.941}

\appendix

\section{Prompts}

\begin{tcolorbox}[colframe=example_front, colback=example_back, coltitle=black, title=\textbf{Inference Prompt Template}]
\textit{\{Input Question\}} + Please reason step by step, and put your final answer within \textbackslash boxed\{\}.
\end{tcolorbox}

\begin{tcolorbox}[breakable, colframe=example_front, colback=example_back, coltitle=black, title=\textbf{CoT Analysis Prompt Template}]
\textbf{System:} You are an expert in analyzing the reasoning processes of Large Language Models. Your task is to segment a raw "Chain of Thought" (CoT) into distinct semantic steps and classify each step.

\vspace{0.5em} \hrule \vspace{0.5em} 

\textbf{\#\#\# Input Data} \\
You will receive:
\begin{enumerate}
    \item The Math Problem.
    \item The raw reasoning text (content within \texttt{\textless think\textgreater} tags).
    \item Whether the problem was solved correctly.
\end{enumerate}

\textbf{\#\#\# Task}
\begin{enumerate}
    \item \textbf{Segmentation}: Break the reasoning text into a list of chronological steps. A "step" is a coherent unit of thought, calculation, or reflection.
    \item \textbf{Classification}: Assign one of the following labels to each step:
    \begin{itemize}
        \item \textbf{Forward}: Constructive reasoning that moves closer to the solution (e.g., proposing a method, performing a calculation, deriving a sub-result).
        \item \textbf{Verification}: Self-correction, double-checking, or validating a previous step (e.g., "Wait, let me check", "Re-calculating this", "Since $17^2$ is 289..."). \textit{Note: This is considered a VALID and useful step.}
        \item \textbf{Redundant}: Unnecessary repetition, getting stuck in a loop, stalled thinking, or restating the obvious without progress (e.g., repeating the question multiple times, "Hmm... let me think", circular logic that adds no value).
    \end{itemize}
\end{enumerate}

\textbf{\#\#\# Output Format (JSON Only)}
\begin{footnotesize}
\begin{texttt}
\{ \\
\hspace*{1em} "total\_steps": 15, \\
\hspace*{1em} "breakdown": \{ \\
\hspace*{2em} "forward\_count": 10, \\
\hspace*{2em} "verification\_count": 3, \\
\hspace*{2em} "redundant\_count": 2 \\
\hspace*{1em} \}, \\
\hspace*{1em} "steps": [ \\
\hspace*{2em} \{"content": "...", "label": "Forward"\}, \\
\hspace*{2em} \{"content": "Wait, let me double check...", "label": "Verification"\}, \\
\hspace*{2em} \{"content": "So the answer is... wait no...", "label": "Redundant"\} \\
\hspace*{1em} ] \\
\}
\end{texttt}
\end{footnotesize}

\end{tcolorbox}

\section{Hyper-parameters}
In this section, we list the hyper-parameters used in different phases of training and inference. We used Llama-Factory~\cite{zheng2024llamafactory} to conduct SFT training (for \emph{DEER+SFT} after rejection sampling), and EasyR1~\cite{zheng2025easyr1} for GRPO training. Table~\ref{tab:hyper} provides the values for the hyper-parameters. The table includes settings such as the number of GPUs, learning rates, batch sizes, and the number of epochs for each phase.

\begin{table}[h!]
    \centering
    \small
    \begin{tabular}{l|c}
    \toprule
    \textbf{Hyper-parameter} & \textbf{Value} \\
    \midrule
    \multicolumn{2}{c}{\textit{\textbf{Supervised Fine-Tuning (SFT)}}} \\
    \midrule
    Number of GPUs (H200) & 8 \\
    Per Device Train Batch Size & 1 \\
    Gradient Accumulation Steps & 2 \\
    Learning Rate & $1.0\times 10^{-5}$ \\
    LR Scheduler & Cosine \\
    Warmup Ratio & 0.1 \\
    Epochs & 3 \\
    \midrule
    \multicolumn{2}{c}{\textit{\textbf{GRPO Training}}} \\
    \midrule
    Number of GPUs (H200) & 8 \\
    Learning Rate & $1.0\times 10^{-6}$ \\
    Weight Decay & $1.0\times 10^{-2}$ \\
    KL Coefficient ($\beta_{KL}$) & 0.01 \\
    Group Size ($G$) & 5 \\
    Rollout Temperature & 1.0 \\
    Global Batch Size & 128 \\
    Rollout Batch Size & 512 \\
    Total Training Steps & 100 \\
    \midrule
    \multicolumn{2}{c}{\textit{\textbf{Inference}}} \\
    \midrule
    Temperature & 0.6 \\
    Top-p & 0.9 \\
    Max Tokens & 32768 \\
    \bottomrule
    \end{tabular}
    \caption{Hyper-parameters for SFT, GRPO training, and Inference.}
    \label{tab:hyper}
\end{table}

\section{Full Reasoning Trajectories}
\begin{tcolorbox}[breakable, colframe=example_front, colback=example_back, coltitle=black, title=\textbf{Demo of Step-GRPO Output}]

\textbf{User Input:} \\
Every morning, Aya does a $9$ kilometer walk, and then finishes at the coffee shop. One day, she walks at $s$ kilometers per hour, and the walk takes $4$ hours, including $t$ minutes at the coffee shop. Another morning, she walks at $s+2$ kilometers per hour, and the walk takes $2$ hours and $24$ minutes, including $t$ minutes at the coffee shop. This morning, if she walks at $s+\frac12$ kilometers per hour, how many minutes will the walk take, including the $t$ minutes at the coffee shop? Please reason step by step, and put your final answer within \textbackslash boxed\{\}.

\vspace{0.5em} \hrule \vspace{0.5em} 

\textbf{Model Prediction:} \\
\textbf{<think>}
{
\small \itshape 
Okay, so I need to figure out how long Aya's walk will take this morning, including the time she spends at the coffee shop. Let me start by understanding the problem step by step.

First, the problem says that every morning Aya does a 9 kilometer walk and then finishes at the coffee shop. So, the total distance she walks each morning is 9 km, right? Then, on one day, she walks at a speed of $s$ km/h, and the walk takes 4 hours, including $t$ minutes at the coffee shop. Another morning, she walks at $s + 2$ km/h, and the walk takes 2 hours and 24 minutes, including $t$ minutes at the coffee shop. Then, this morning, she walks at $s + 1/2$ km/h, and we need to find how many minutes the walk takes, including $t$ minutes at the coffee shop.

Hmm, okay. Let me parse this again. The key here is that the total time includes both the walking time and the time spent at the coffee shop. So, when they say ``the walk takes 4 hours, including $t$ minutes at the coffee shop,'' that means the total time from starting to finish at the coffee shop is 4 hours, which includes both the walking time and the $t$ minutes at the coffee shop. Similarly for the other day.

So, let me denote:
For the first day: Total time = walking time + $t$ minutes = 4 hours. But walking time is distance divided by speed. The distance is 9 km, speed is $s$ km/h. So walking time is $9/s$ hours. Then, total time is $9/s + t/60$ hours = 4 hours.

Similarly, for the second day: Total time is 2 hours and 24 minutes, which is $2 + 24/60 = 2.4$ hours. So, same as before, walking time is $9/(s + 2)$ hours, and total time is $9/(s + 2) + t/60 = 2.4$ hours.

So, we have two equations:
1) $9/s + t/60 = 4$
2) $9/(s + 2) + t/60 = 2.4$

We need to solve these equations to find $s$ and $t$, then use them to find the total time for the third day, which is walking time at $s + 1/2$ km/h plus $t$ minutes.

So, first, let me write equations 1 and 2.
Equation 1: $9/s + t/60 = 4$. Equation 2: $9/(s + 2) + t/60 = 2.4$.
So, if I subtract equation 2 from equation 1, the $t/60$ terms will cancel out. Let me try that.
Equation 1 - Equation 2: $[9/s - 9/(s + 2)] + [t/60 - t/60] = 4 - 2.4$.
Simplify: $9/s - 9/(s + 2) = 1.6$.

So, compute $9/s - 9/(s + 2) = 1.6$. Let me compute the left side:
$9[1/s - 1/(s + 2)] = 9[(s + 2 - s)/[s(s + 2)]] = 9[2 / (s(s + 2))] = 18 / [s(s + 2)]$.
So, $18 / [s(s + 2)] = 1.6$. Let me write 1.6 as $8/5$, so: $18 / [s(s + 2)] = 8/5$.

Cross multiplying: $18 \times 5 = 8 \times s(s + 2) \Rightarrow 90 = 8s^2 + 16s$.
Bring all terms to one side: $8s^2 + 16s - 90 = 0$.
Divide both sides by 2: $4s^2 + 8s - 45 = 0$.

Now, solving quadratic equation $4s^2 + 8s - 45 = 0$. Using quadratic formula: $s = [-b \pm \sqrt{b^2 - 4ac}] / 2a$.
Here, $a = 4, b = 8, c = -45$.
Discriminant $D = 8^2 - 4 \times 4 \times (-45) = 64 + 720 = 784$. $\sqrt{784} = 28$.

So, $s = (-8 \pm 28) / 8$.
We have two solutions: $s = (-8 + 28)/8 = 20/8 = 2.5$ and $s = (-8 - 28)/8 = -36/8 = -4.5$.
But speed can't be negative, so $s = 2.5$ km/h.

Now, let me find $t$ from equation 1: $9/s + t/60 = 4$.
So, $9 / 2.5 + t / 60 = 4$. Compute $9 / 2.5$: 9 divided by 2.5 is same as $9 \times 2 / 5 = 18 / 5 = 3.6$ hours.
So, $3.6 + t / 60 = 4 \Rightarrow t / 60 = 4 - 3.6 = 0.4$ hours.
Convert 0.4 hours to minutes: $0.4 \times 60 = 24$ minutes. So, $t = 24$ minutes.

So, now, the problem says: This morning, if she walks at $s + 1/2$ km/h, how many minutes will the walk take, including the $t$ minutes at the coffee shop?
So, $s$ is 2.5, so $s + 1/2$ is $2.5 + 0.5 = 3$ km/h.
So, her speed is 3 km/h. The distance is still 9 km. So, walking time is $9 / 3 = 3$ hours.
Then, total time is walking time + $t$ minutes = 3 hours + 24 minutes.
Convert 3 hours to minutes: $3 \times 60 = 180$ minutes. Then add 24 minutes: $180 + 24 = 204$ minutes.

Therefore, the total time is 204 minutes.
Let me check if this is consistent with the second day's data.
On the second day, she walked at $s + 2 = 2.5 + 2 = 4.5$ km/h. Walking time is $9 / 4.5 = 2$ hours. Then total time is 2 hours + 24 minutes = 2 hours 24 minutes, which is 2.4 hours, which matches the given data. So that's correct.

So, the answer should be 204 minutes.
}
\textbf{</think>}

\vspace{0.5em}

To determine how long Aya's walk takes this morning, including the time spent at the coffee shop, we need to analyze the relationship between her walking speed, the distance, and the total time (including the fixed time at the coffee shop).

\textbf{Step 1: Define the Variables} \\
Let $s$ be Aya's walking speed in km/h and $t$ be the fixed time in minutes at the coffee shop.
We are given:
\begin{itemize}
    \item \textbf{Day 1:} Speed $s$, Total time 4h.
    \begin{equation} \frac{9}{s} + \frac{t}{60} = 4 \tag{1} \label{eq:1} \end{equation}
    \item \textbf{Day 2:} Speed $s+2$, Total time 2.4h.
    \begin{equation} \frac{9}{s+2} + \frac{t}{60} = 2.4 \tag{2} \label{eq:2} \end{equation}
\end{itemize}

\textbf{Step 2: Eliminate $t$ to Solve for $s$} \\
Subtract equation (\ref{eq:2}) from equation (\ref{eq:1}):
\begin{align*}
\left( \frac{9}{s} - \frac{9}{s + 2} \right) &= 4 - 2.4 = 1.6 \\
9 \left( \frac{s + 2 - s}{s(s + 2)} \right) &= 1.6 \\
\frac{18}{s(s + 2)} &= \frac{8}{5} \\
90 &= 8s^2 + 16s \\
4s^2 + 8s - 45 &= 0
\end{align*}

Use the quadratic formula:
\begin{align*}
s &= \frac{-8 \pm \sqrt{8^2 - 4(4)(-45)}}{2(4)} \\
  &= \frac{-8 \pm \sqrt{64 + 720}}{8} \\
  &= \frac{-8 \pm 28}{8}
\end{align*}
We get $s = 2.5$ (discarding $s = -4.5$).

\textbf{Step 3: Solve for $t$} \\
Plug $s = 2.5$ into equation (\ref{eq:1}):
\begin{align*}
\frac{9}{2.5} + \frac{t}{60} &= 4 \\
3.6 + \frac{t}{60} &= 4 \quad \Rightarrow \quad t = 24 \text{ minutes}
\end{align*}

\textbf{Step 4: Compute Total Time This Morning} \\
This morning, Aya walks at $s + 0.5 = 3$ km/h.
\begin{align*}
\text{Total Time} &= \frac{9}{3} \text{ hours} + 24 \text{ minutes} \\
&= 3 \times 60 + 24 \\
&= 180 + 24 = 204 \text{ minutes}
\end{align*}

\textbf{Final Answer}
\[ \boxed{204} \]

\end{tcolorbox}

\section{Generalization Capabilities}
To verify the broad universality of Step-GRPO and rule out the possibility that our method overfits to specific models or domains, we evaluate its performance across two crucial dimensions: differing model architectures and non-mathematical reasoning domains.

\paragraph{Generalization to Other Architectures} We conducted supplementary experiments substituting the base model with the DeepSeek-R1-Distill-Llama-8B architecture. The experimental results show that Step-GRPO also demonstrates robust performance on the Llama architecture. Specifically, Step-GRPO achieved the best overall token compression rate (CR = 66.79\%) compared to Vanilla (100.00\%), GRPO (109.03\%), and GRPO-$\lambda$ (81.83\%) while maintaining a highly competitive overall accuracy of 65.40\%. 

\begin{table*}[t]
\centering
\caption{Generalization results on DeepSeek-R1-Distill-Llama-8B architecture.}
\label{tab:generalization}
\resizebox{\linewidth}{!}{
\begin{tabular}{l|cc|cc|cc|cc|cc|cc|cc}
\toprule
\multirow{2}{*}{\textbf{Method}} & \multicolumn{2}{c|}{\textbf{GSM8K}} & \multicolumn{2}{c|}{\textbf{GPQA Diamond}} & \multicolumn{2}{c|}{\textbf{MATH-500}} & \multicolumn{2}{c|}{\textbf{AMC 23}} & \multicolumn{2}{c|}{\textbf{AIME 2024}} & \multicolumn{2}{c|}{\textbf{AIME 2025}} & \multicolumn{2}{c}{\textbf{Overall}} \\
 & Acc & Tok & Acc & Tok & Acc & Tok & Acc & Tok & Acc & Tok & Acc & Tok & Acc & \textbf{CR} \\
\midrule
Vanilla & 87.95 & 1673.61 & 46.97 & 8919.74 & 89.80 & 4237.33 & 87.50 & 6151.73 & 46.67 & 15542.40 & 23.33 & 14925.97 & 63.70 & 100.00 \\
GRPO & 89.23 & 2228.22 & 50.51 & 9084.33 & 93.20 & 4476.27 & 90.00 & 7792.75 & 53.33 & 13407.43 & 26.67 & 15017.83 & 67.20 & 109.03 \\
GRPO-$\lambda$ & 86.96 & 987.39 & 52.02 & 8082.42 & 90.40 & 3138.21 & 87.50 & 5633.93 & 46.67 & 12711.67 & 30.00 & 14019.73 & 65.60 & 81.83 \\
\textbf{Step-GRPO (Ours)} & 87.49 & 976.08 & 51.01 & 6984.80 & 88.60 & 2710.77 & 85.00 & 5060.15 & 50.00 & 9105.30 & 30.00 & 8848.23 & 65.40 & \textbf{66.79} \\
\bottomrule
\end{tabular}
}
\end{table*}

On highly challenging tasks like AIME and relatively difficult tasks such as GPQA, Step-GRPO achieves significant token compression while maintaining or even improving accuracy. This indicates that our formulation—optimizing relative semantic efficiency—effectively adapts to reasoning models with diverse, idiosyncratic output characteristics without relying on model-specific features.

\paragraph{Generalization to Non-Mathematical Domains} To further demonstrate that our approach transfers beyond domain-specific math problems, we evaluated Step-GRPO on BIG-Bench Hard (BBH) ~\citep{suzgun2023challenging}, a benchmark emphasizing symbolic and abstract logical reasoning. As shown in Table~\ref{tab:bbh_result}, Step-GRPO achieves the highest accuracy ($86.51\%$) across all methods while maintaining a superior compression rate compared to standard GRPO and SFT baselines. This confirms that the internalized cognitive state transitions captured by our linguistic triggers apply universally across diverse reasoning paradigms.

\begin{table}[h]
\centering
\caption{Performance on the BIG-Bench Hard (BBH) benchmark.}
\label{tab:bbh_result}
\resizebox{\linewidth}{!}{
\begin{tabular}{l|cc|c}
\toprule
\textbf{Method} & \textbf{Acc} & \textbf{Tokens} & \textbf{CR} \\
\midrule
Vanilla & 86.14 & 1337 & 100.0 \\
GRPO & 85.96 & 1247 & 93.2 \\
DEER-SFT & 84.89 & 1520 & 113.6 \\
GRPO-$\lambda$ & 86.30 & 989 & 74.0 \\
GRPO-LP & 84.81 & 811 & 60.7 \\
GRPO-SOP & 86.46 & 1084 & 81.1 \\
\textbf{Step-GRPO (Ours)} & \textbf{86.51} & 1097 & \textbf{82.1} \\
\bottomrule
\end{tabular}
}
\end{table}

\section{Ablation on Confidence Threshold}
The dynamic truncated rollout relies on the threshold $\delta$ to decide when to pause generation. A potential concern is whether confidence estimates correlate perfectly with correctness, as models can be ``confidently wrong'' on harder instances. However, Step-GRPO does not rely solely on the pre-trained priors; the RL process itself serves as a continuous calibration mechanism. During training, trajectories that are ``short, high-confidence, but factually incorrect'' receive severe negative rewards. This dense error-signal forces the model to gradually align its confidence probability with actual correctness. To demonstrate the robustness of our dynamic exit criterion, we performed an ablation on the threshold parameter $\delta$. The overall results in Table~\ref{tab:threshold} show that while lower thresholds ($\delta = 0.80$, $\delta = 0.90$) induce more aggressive truncation leading to faster performance drops, our chosen threshold ($\delta = 0.95$) effectively balances accuracy and token usage. Notably, the method comfortably outperforms baselines across a broad, reasonable range of confidence thresholds, reinforcing the validity of the stopping criterion.

\begin{table*}[t]
\centering
\caption{Ablation on confidence threshold $\delta$. We report Accuracy (\%) across six benchmarks and Overall Compression Rate (CR).}
\label{tab:threshold}
\resizebox{\linewidth}{!}{
\begin{tabular}{l|cccccc|cc}
\toprule
\textbf{Model / Threshold} & \textbf{GSM8K} & \textbf{GPQA} & \textbf{MATH-500} & \textbf{AMC 23} & \textbf{AIME 24} & \textbf{AIME 25} & \textbf{Overall (Acc)} & \textbf{Overall (CR)} \\
\midrule
Vanilla & 93.2 & 62.1 & 96.8 & 97.5 & 70.0 & 60.0 & 79.9 & 100.0 \\
\midrule
$\delta = 0.80$ & 93.3 & 55.6 & 94.0 & 92.5 & 66.7 & 46.7 & 74.8 & 62.0 \\
$\delta = 0.90$ & 93.4 & 56.1 & 95.4 & 90.0 & 70.0 & 56.7 & 76.9 & 64.0 \\
\textbf{Step-GRPO ($\delta = 0.95$)} & 94.9 & 56.1 & 96.8 & 95.0 & 76.7 & 73.3 & \textbf{82.1} & 68.0 \\
\bottomrule
\end{tabular}
}
\end{table*}

\section{Empirical Stability Evaluation}
To further address variance and confirm stability, we conducted independent runs across all benchmarks on Qwen3-8B. Table~\ref{tab:stability} reports the Accuracy Mean $\pm$ Standard Deviation across three random seeds. Step-GRPO achieves the highest overall stability, consistently minimizing outcome variance effectively. In contrast, static length-penalty methods demonstrate significant instability, especially on highly demanding tasks such as AIME.

\begin{table*}[t]
\centering
\caption{Accuracy (Mean $\pm$ Std) across 3 independent runs on Qwen3-8B.}
\label{tab:stability}
\resizebox{\linewidth}{!}{
\begin{tabular}{l|cccccc|c}
\toprule
\textbf{Method} & \textbf{GSM8K} & \textbf{MATH-500} & \textbf{GPQA} & \textbf{AMC 23} & \textbf{AIME 24} & \textbf{AIME 25} & \textbf{Overall} \\
\midrule
Vanilla & 93.1 $\pm$ 0.4 & 96.5 $\pm$ 0.3 & 57.1 $\pm$ 0.0 & 95.0 $\pm$ 0.0 & 67.8 $\pm$ 5.1 & 62.2 $\pm$ 7.7 & 78.6 $\pm$ 2.3 \\
DEER+SFT & 90.8 $\pm$ 0.3 & 94.4 $\pm$ 0.7 & 46.8 $\pm$ 0.6 & 87.5 $\pm$ 2.5 & 57.8 $\pm$ 3.9 & 51.1 $\pm$ 3.9 & 71.4 $\pm$ 2.0 \\
GRPO & 93.3 $\pm$ 0.3 & 96.4 $\pm$ 0.2 & 57.7 $\pm$ 1.5 & 93.3 $\pm$ 1.4 & 76.7 $\pm$ 3.3 & 58.9 $\pm$ 6.9 & 79.4 $\pm$ 2.3 \\
GRPO-$\lambda$ & 92.9 $\pm$ 0.2 & 96.1 $\pm$ 0.3 & 57.1 $\pm$ 2.2 & 92.5 $\pm$ 0.0 & 73.3 $\pm$ 3.3 & 54.5 $\pm$ 6.9 & 77.7 $\pm$ 2.2 \\
GRPO+LP & 93.7 $\pm$ 0.3 & 95.1 $\pm$ 0.2 & 58.3 $\pm$ 1.8 & 93.3 $\pm$ 2.9 & 68.9 $\pm$ 3.9 & 50.0 $\pm$ 5.8 & 76.5 $\pm$ 2.5 \\
GRPO+SOP & 92.5 $\pm$ 0.2 & 96.5 $\pm$ 0.1 & 57.2 $\pm$ 1.9 & 92.5 $\pm$ 0.0 & 75.6 $\pm$ 1.9 & 62.2 $\pm$ 5.1 & 79.4 $\pm$ 1.6 \\
Step-GRPO & \textbf{94.8} $\pm$ 0.0 & \textbf{96.5} $\pm$ 0.2 & 56.6 $\pm$ 0.9 & \textbf{95.8} $\pm$ 1.4 & \textbf{77.8} $\pm$ 5.1 & \textbf{68.9} $\pm$ 3.9 & \textbf{81.7} $\pm$ 1.9 \\
\bottomrule
\end{tabular}
}
\end{table*}

\section{Training Overhead}
To clearly demonstrate the trade-off in computational cost introduced by the Dynamic Truncated Rollout, we compare the average per-step time overhead (in seconds) of standard GRPO and Step-GRPO on the Qwen3-8B model. As shown in Table~\ref{tab:overhead}, we evaluate both the initial phase (first 50 steps) and the convergence phase (last 50 steps).

\begin{table}[ht]
\centering
\caption{Average per-step time overhead (seconds) on Qwen3-8B.}
\label{tab:overhead}
\resizebox{\linewidth}{!}{
\begin{tabular}{l|ccccc}
\toprule
\textbf{Method} & \textbf{Generation} & \textbf{Update} & \textbf{Old Policy} & \textbf{Ref Model} & \textbf{Reward} \\
\midrule
GRPO (0-50 steps) & 1184.8 & 928.5 & 183.4 & 187.1 & 3.8 \\
Step-GRPO (0-50 steps) & 1678.1 & 659.2 & 101.3 & 101.1 & 3.8 \\
\midrule
GRPO (50-100 steps) & 1111.6 & 904.1 & 179.4 & 183.1 & 3.6 \\
Step-GRPO (50-100 steps) & 1392.7 & 551.3 & 85.8 & 85.6 & 3.8 \\
\bottomrule
\end{tabular}
}
\end{table}

While the \emph{Generation} phase naturally incurs an initial overhead due to the dynamic confidence estimation and answer induction during rollout, this is effectively compensated for by significant time savings in the \emph{Update}, \emph{Old Policy}, and \emph{Reference Model} phases. This reduction occurs because as the model rapidly learns to generate shorter, more efficient sequences, the computational burden for backward passes and likelihood calculations drops dramatically. Notably, as the model converges (Steps 50-100), Step-GRPO actually becomes faster overall due to these escalating efficiency gains, demonstrating that the initial rollout calculation is a highly worthwhile investment for an efficient training loop.

\section{Robustness of Heuristic Triggers}
A natural concern regarding our methodology is the sensitivity of the heuristic trigger words (\emph{e.g.}, ``Wait'', ``Alternatively'') used to segment semantic steps. While exploring parameter-free or learned segmentation models represents a promising future direction, we argue that our current approach provides sufficient robustness for two primary reasons. First, \emph{Action Transition Points}: recent literature~\citep{yang2025dynamic,qiao2025concise} validates these markers as Action Transition Points that naturally signal a high-entropy cognitive shift from linear deduction to self-verification or branch correction. They serve as semantic anchors that effectively delineate valid reasoning steps from redundant loops, demonstrating that they are fundamental to multi-step reasoning processes rather than artifacts of a specific writing style or language. Second, \emph{RL Adaptability and Invariance}: while heuristic, these triggers serve as initial anchors for an adaptive RL process. Step-GRPO fundamentally relies on the group-relative baseline (Equation 4). Even if the chosen trigger set is imperfect—such as missing occasional boundary markers in a specific domain—the resulting ``step count bias'' is uniformly applied across all completions within the same sample group $\mathcal{G}$, ensuring the \emph{relative ranking} of generation efficiency remains consistent and valid. Furthermore, the model learns to align its internal confidence with these structural boundaries to maximize the reward. This structural resilience allows the RL gradient to correctly favor shorter valid paths, making the system inherently robust to minor linguistic variations and the absolute granularity defining a single ``step''.

\end{document}